%% file: main.tex
\documentclass{article}

\usepackage[preprint]{neurips_2026}

\usepackage[utf8]{inputenc} 
\usepackage[T1]{fontenc}    
\usepackage{hyperref}       
\usepackage{url}            
\usepackage{booktabs}       
\usepackage{amsfonts}       
\usepackage{nicefrac}       
\usepackage{microtype}      
\usepackage{xcolor}         
\usepackage{graphicx}
\usepackage{cleveref}
\usepackage[table]{xcolor}

\title{Cross-Market Meta-Learning for Electricity Price Forecasting%
\thanks{Accepted at the NeurIPS 2026 Workshop on AI Foundations for Power Grids.}}
\workshoptitle{AI Foundations for Power Grids}

\author{%
  Delaram Pirhayatifard \\
  Rice University\\
  Houston, TX \\
  \texttt{dp43@rice.edu}
  \And
  Terrence Alsup \\
  Morgan Stanley \\
  New York, NY \\
  \texttt{Terrence.Alsup@morganstanley.com}
}

\begin{document}

\maketitle

\begin{abstract} Day-ahead (DA) electricity price forecasting is usually posed per market, with a separate model fitted to each grid's own history. This caps accuracy: statistical and tree-based baselines saturate, while higher-capacity sequence models need more training data than a single market's few years of hourly prices can supply. Such methods also treat each node as an independent series, discarding the spatial dependencies induced by network topology and congestion. Because price formation across independent system operators (ISOs) is driven by shared physics and largely common market design, we instead treat cross-market history as a pretraining corpus. We present MetaGrid, a topology-aware transformer, meta-trained across markets so that a single zone-count-independent parameter set adapts to an unseen grid. On a leak-free benchmark of five U.S. ISOs, where every model receives the same prices, DA load forecasts and calendar features, MetaGrid attains the lowest average MAE, improving on the best baseline by 3.1\% and on the same architecture trained from scratch by 7.7\%. 
\end{abstract}

\section{Introduction}
\label{sec:intro}
\input{introduction}

\section{Problem Setup \& Methodology}
\label{sec:method}

\input{method_final}
\section{Experiments}
\textbf{Baselines.} We compare against naive forecasts (persistence 24h, 48h), regression/GBM models (XGBoost, LightGBM, quantile regression, LEAR), a tabular foundation model (TabPFN), graph baseline (MTGNN), a time-series model (PatchTST), and a time-series foundation model (Chronos T5-large, adapted). All models receive the same inputs, trailing prices, the day-ahead load forecast and calendar features, and are run with five seeds varying initialization, meta-episode sampling and evaluation-time adaptation. Splits are chronological and identical for every model and ISO: training through 2023-12-31, 2024 for validation, 2025 held out for test. More details on data and implementation in Appendix~\cref{app:data,app:impl}. \footnote{Code is available at:
\url{https://github.com/delaramphf/MetaGrid.git}. Load and price data are downloaded from that ISO's own market data
portal.}

\textbf{Metrics.} Point accuracy via MAE and probabilistic metric AQL (average quantile loss). 

\begin{figure}[t]
    \centering
    \begin{minipage}{0.5\linewidth}
        \centering
        \includegraphics[width=\linewidth]{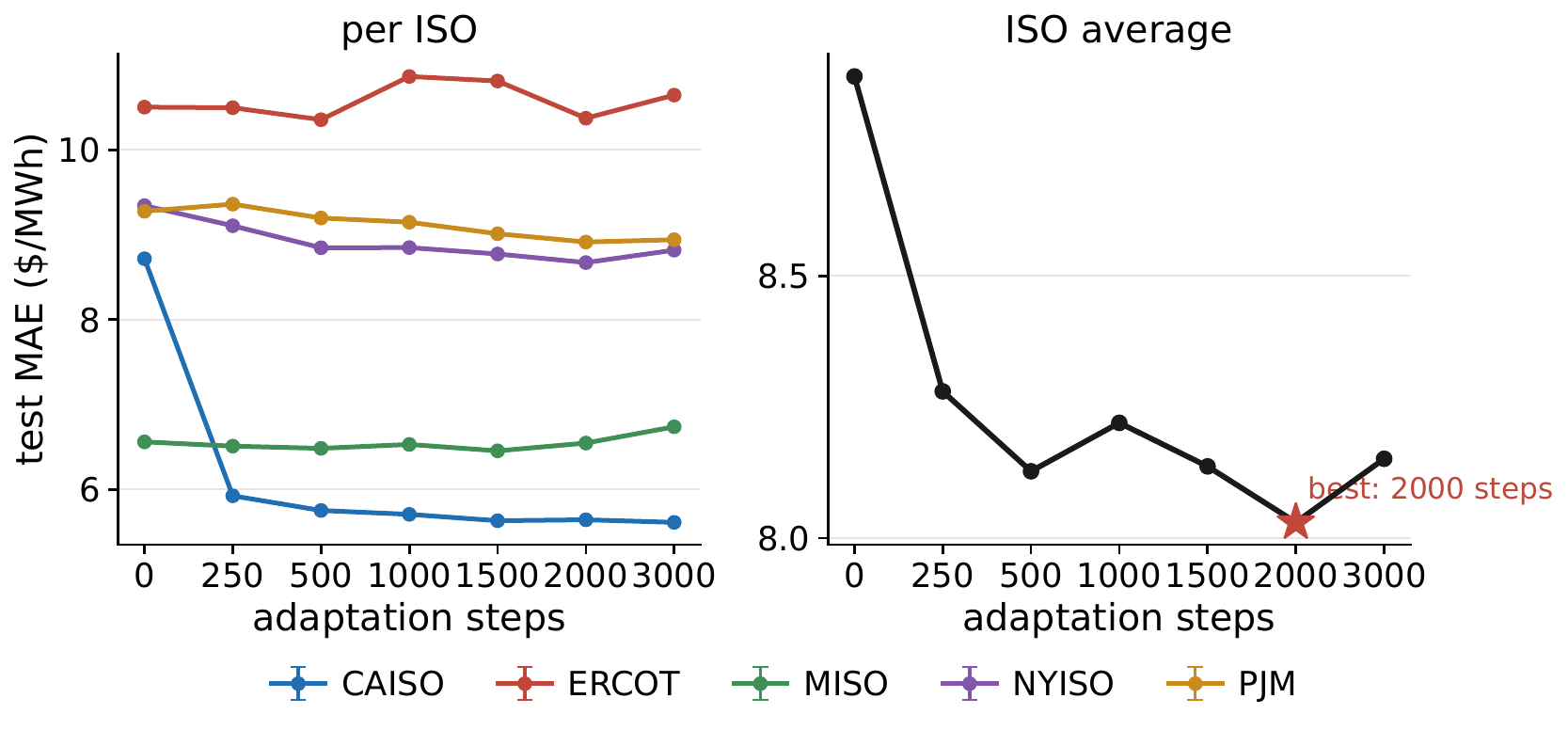}
        \caption{Test MAE vs. adaptation steps on the held-out target, per ISO (left) and averaged (right). Zero steps is zero-shot transfer. }
        \label{fig:adaptation_curve}
    \end{minipage}
    \hfill
    \begin{minipage}{0.45\linewidth}
        \centering
        \includegraphics[width=\linewidth]{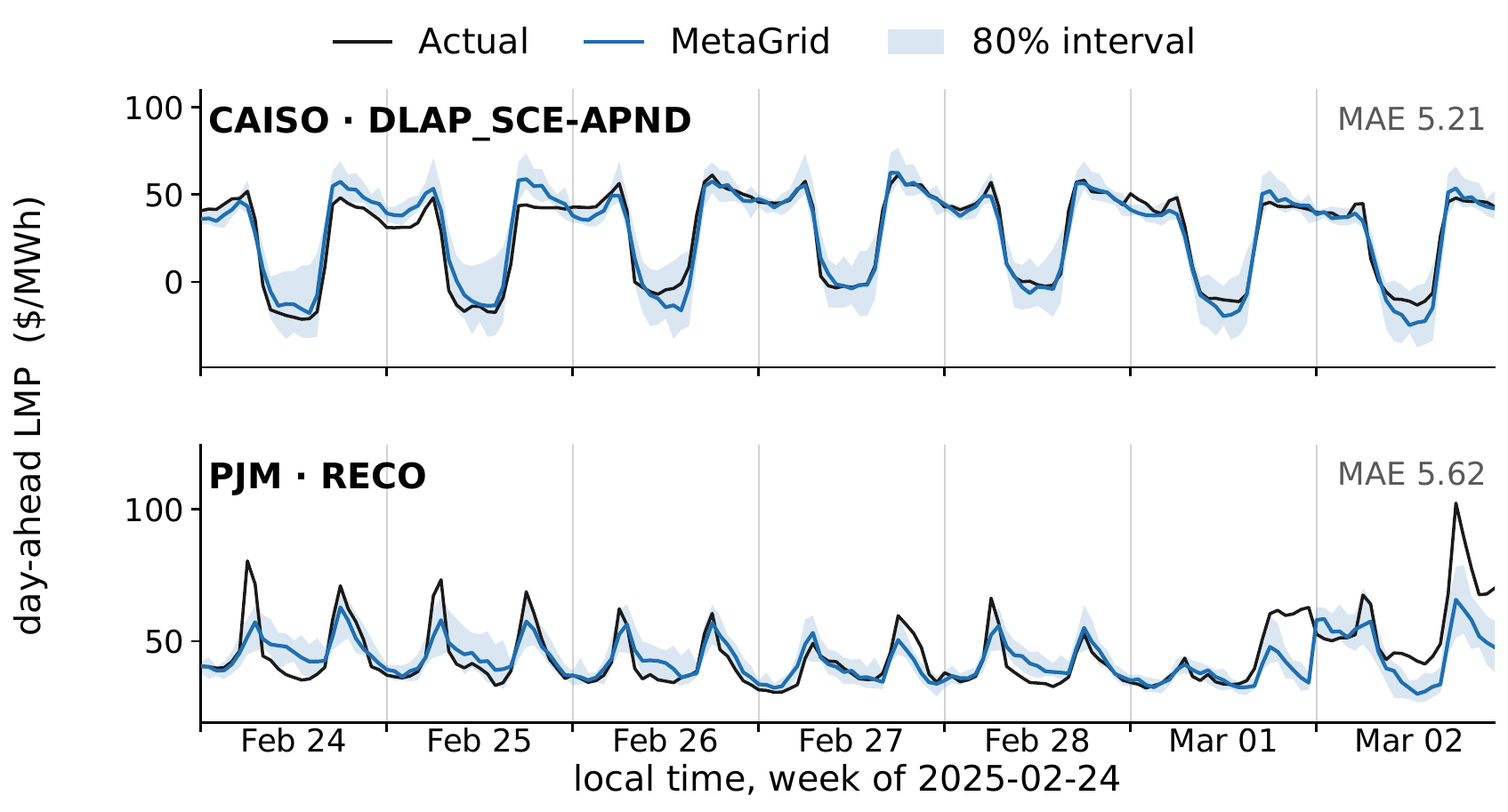}
        \caption{One week of DA LMP test forecast against realized LMP; the band is the 10-90\% quantile.}
        \label{fig:forecast_fig}
    \end{minipage}
\end{figure}

\textbf{Performance on held-out target ISO against the baselines}
\label{sec:full_baselines}
Table~\ref{tab:baseline_full} reports our model against every baseline category we ran at full
adaptation data, averaged over all five ISOs. Our model achieves the best MAE and AQL of any model in this comparison,
while using fewer parameters than any competing neural model. MetaGrid uses 7.8M parameters which is within 2× of PatchTST and roughly 90× smaller than Chronos-T5-large, yet attains 3.1\% lower MAE and 3.5\% lower average quantile loss than the strongest baseline. MetaGrid wins on 4 out of 5 ISOs, except for ERCOT, where TSFM baselines outperform MetaGrid. Figure~\ref{fig:forecast_fig} illustrates one week of DA forecast for CAISO and PJM compared to the ground truth. MetaGrid tracks the ground-truth
time series at most time steps $t$. 

The gain is not simply due to model scale: removing the graph-aware cross-zone module increases MAE by 0.39 \$/MWh and degrades performance on all five ISOs (Appendix D). Also, since $C_{i,o}$ is recomputed at each origin while $A_i$ is fixed, the
zone relationships the model uses change over time instead of staying fixed,
and the learned weights differ by market: NYISO, with persistent congestion,
relies more on adjacency ($\alpha = 0.71$) while MISO leans on correlation
($\beta = 0.54$, Table~\ref{tab:priors}).

\begin{table*}[t]
\centering
\caption{Per-ISO forecasting performance on the 2025 test set (5 seeds).
$\dagger$ denotes deterministic models; bold indicates the best result.}
\label{tab:baseline_full}
\resizebox{\textwidth}{!}{%
\begin{tabular}{llcccccc|c}
\toprule
 & & \multicolumn{6}{c|}{MAE (\$/MWh)} & AQL \\
Category & Model & PJM & NYISO & CAISO & MISO & ERCOT & Avg & (\$/MWh) \\
\midrule
Naive & Persistence (24h)$^{\dagger}$ & $11.61$ & $11.07$ & $6.65$ & $8.11$ & $12.90$ & $10.068$ & -- \\
Naive & Persistence (48h)$^{\dagger}$ & $16.15$ & $16.41$ & $9.19$ & $10.78$ & $15.66$ & $13.639$ & -- \\
Linear & LEAR$^{\dagger}$ & $9.24$ & $9.09$ & $6.34$ & $7.64$ & $11.07$ & $8.675$ & $3.544$ \\
Linear & Quantile regression & $11.00_{\pm 0.01}$ & $9.94_{\pm 0.01}$ & $6.75_{\pm 0.02}$ & $7.69_{\pm 0.04}$ & $13.79_{\pm 0.02}$ & $9.833_{\pm 0.010}$ & $4.009_{\pm 0.002}$ \\
GBM & XGBoost & $10.09_{\pm 0.03}$ & $9.96_{\pm 0.06}$ & $6.27_{\pm 0.01}$ & $7.90_{\pm 0.05}$ & $11.19_{\pm 0.06}$ & $9.080_{\pm 0.018}$ & $3.757_{\pm 0.007}$ \\
GBM & LightGBM & $9.77_{\pm 0.09}$ & $10.02_{\pm 0.07}$ & $6.09_{\pm 0.02}$ & $7.89_{\pm 0.02}$ & $11.32_{\pm 0.22}$ & $9.016_{\pm 0.046}$ & $3.684_{\pm 0.005}$ \\
Tabular FM & TabPFN & $9.19_{\pm 0.08}$ & $8.99_{\pm 0.10}$ & $5.90_{\pm 0.03}$ & $7.94_{\pm 0.10}$ & $10.28_{\pm 0.02}$ & $8.457_{\pm 0.013}$ & $3.391_{\pm 0.007}$ \\
Graph & MTGNN & $10.46_{\pm 0.004}$ & $13.87_{\pm 0.2}$ & $7.97_{\pm 0.01}$ & $8.01_{\pm 0.08}$ & $15.92_{\pm 0.03}$ & $11.26_{\pm 0.004}$ & $4.625_{\pm 0.003}$ \\
Time-series & PatchTST & $9.24_{\pm 0.03}$ & $8.91_{\pm 0.05}$ & $5.89_{\pm 0.01}$ & $7.03_{\pm 0.05}$ & $10.59_{\pm 0.03}$ & $8.333_{\pm 0.014}$ & $3.380_{\pm 0.006}$ \\
Foundation & Chronos T5-large & $9.53_{\pm 0.003}$ & $9.46_{\pm 0.005}$ & $6.10_{\pm 0.003}$ & $6.58_{\pm 0.012}$ & $\mathbf{9.77_{\pm 0.009}}$ & $8.289_{\pm 0.011}$ & $3.427_{\pm 0.008}$ \\
\midrule
Ours & MetaGrid (scratch) & $9.11_{\pm 0.07}$ & $9.18_{\pm 0.12}$ & $5.71_{\pm 0.03}$ & $8.37_{\pm 0.07}$ & $11.12_{\pm 0.26}$ & $8.699_{\pm 0.062}$ & $3.652_{\pm 0.005}$ \\
\rowcolor{gray!20}
Ours & MetaGrid & $\mathbf{9.04}_{\pm 0.009}$ & $\mathbf{8.78}_{\pm 0.06}$ & $\mathbf{5.64}_{\pm 0.06}$ & $\mathbf{6.43}_{\pm 0.005}$ & $10.27_{\pm 0.004}$ & $\mathbf{8.032}_{\pm 0.028}$ & $\mathbf{3.263}_{\pm 0.002}$ \\
\bottomrule
\end{tabular}%
}
\end{table*}

\textbf{Pretrained vs.\ scratch: effect of adaptation data}
\label{sec:main_results}
Figure~\ref{fig:adaptation_curve} varies adaptation from 0 to 2000 steps, with the target-only model as reference and the optimal step selected on validation. Zero-shot transfer is already competitive in some markets: MISO achieves $\$6.43$/MWh without seeing MISO prices during training. Overall, pretraining is most effective when combined with adaptation.

\input{conclusion}

\bibliographystyle{plain}
\bibliography{ref}

\input{appendix_final}

\end{document}

%% file: introduction.tex
\label{sec:intro}

Wholesale electricity markets clear a day-ahead auction that sets locational marginal prices (LMPs) for every hour of the following delivery day. Because bids must be submitted before the auction closes, participants commit to positions without knowing the prices at which they will settle, and a forecast of the next day's LMPs is what those bidding decisions are made against.

The dominant modeling practice is per-market and
per-zone: a model is fitted to one grid's own history, and each zone is treated as an
independent series~\cite{weron2014electricity,lago2021forecasting}. Two limitations follow. First,
the available sample is small; a single ISO offers a few years of hourly prices, which is
enough to saturate gradient-boosted trees~\cite{chen2016xgboost,ke2017lightgbm} but not
enough to train a high-capacity sequence model~\cite{nie2022time}.
Second, treating zones independently ignores the physical constraints of a real grid, namely congestion, which can be supplied to the model as a prior rather than left to be learned from the target market's history alone~\cite{kipf2016semi,velivckovic2017graph,owerko2020optimal}. Cross-market transfer is known to help single-series price
forecasting~\cite{gunduz2023transfer}; the multi-zone
setting, where markets differ in zone count and topology, is less
well studied.

Both limitations point in the same direction. Price formation across ISOs is governed by the
same physics and largely common market design, so history from other markets isn't
out-of-distribution data to be discarded but a pretraining corpus, the premise underlying
time-series foundation models~\cite{ansari2024chronos,woo2024unified}. This raises the question that organizes this paper: what has to be shared across markets, and what has to remain market-specific? Grids differ in topology, generation mix and scarcity behavior, so the transferable part of a model must not depend on shape of graph it was trained on.

We propose a model that meta-trains across markets so that adaptation to a held-out grid is
itself the training objective~\cite{finn2017model,nichol2018first,toghani2023first},
with every parameter shared across markets and independent of zone count, so the model transfers to a grid it has never seen. The same view has been applied at test
time on graph-structured data~\cite{pirhayatifard2025cross}, where the deployment graph differs
from the training graph. Shared-backbone models with task-specific fine-tuning have also been explored for
probabilistic forecasting with diffusion decoders~\cite{pirhayatifard2023improving}.

Our architecture, MetaGrid, is a patch-based
transformer~\cite{nie2022time} extended with attention across zones, so that
each zone's representation is formed jointly with the others rather than in isolation. Grid structure enters this attention as a soft, learned prior: adjacency and context correlation are weighted by bounded coefficients learned during meta-training, leaving the model free to use the topology or ignore it. The model predicts quantiles and is trained with the pinball
loss~\cite{koenker1978regression,salinas2020deepar}.
We evaluate on a benchmark spanning five U.S. ISOs. Every model sees only information available before that market's day-ahead gate closes, and all are scored on same set of hours against raw published prices.
Against ten diverse baselines~\cite{chen2016xgboost,ke2017lightgbm,hollmann2022tabpfn,ansari2024chronos,wu2020connecting}, MetaGrid attains the lowest average MAE, improving on the strongest baseline by 3.1\% and on the same architecture trained from
scratch by 7.7\%.

Our contributions are: (i) a five-market benchmark with leakage-free evaluation across ten baselines; (ii) MetaGrid, a meta-trained topology-aware transformer whose every parameter is independent of zone count, so a single initialization applies to an unseen grid of any size; and (iii) evidence that transfer improves accuracy on an unseen grid: on every held-out ISO, pretraining beats target-only training of the same architecture.

%% file: method_final.tex
\paragraph{Problem setup}

We consider independent system operators (ISOs) indexed by $i\in\mathcal{I}$.
ISO $i$ is a graph $\mathcal{G}_i=(\mathcal{V}_i,\mathcal{E}_i)$ with
pricing zones $\mathcal{V}_i$, and zone relationships
$\mathcal{E}_i$. We represent $\mathcal{E}_i$ by its adjacency matrix $A_i$.
The number of zones $n_i=|\mathcal{V}_i|$ may differ across ISOs. The graph is
built from market-map proximity and is a coarse structural prior.

For zone $v\in\mathcal{V}_i$ and time $t$, let $y_{i,v,t}$ denote the
day-ahead price. At forecast issue time $o$ the model receives a price
history, the day-ahead load forecast published for the delivery horizon, and calendar features for the
context and delivery horizon; we denote this leak-free information by
$\mathcal{F}_{i,o}$ (Appendix~\ref{app:method}). Given a set of quantile levels $\mathcal{T}$ containing
the median, the model estimates
\begin{equation}
  \widehat q_{i,v,h}^{(\tau)}
  \approx Q_\tau\!\left(y_{i,v,o+h}\mid\mathcal{F}_{i,o}\right),
  \qquad
  h=1,\ldots,H,\quad \tau\in\mathcal{T}.
  \label{eq:forecast_task}
\end{equation}

\subsection{Shared temporal representation}

Price levels and volatility differ widely across markets and zones. We apply reversible instance normalization~\cite{kim2021reversible} per zone over the trailing context, inverted exactly after decoding, so the encoder sees shape rather than level. Within each zone the normalized inputs are divided into nonoverlapping
temporal patches and mapped to tokens by a shared linear projection. A
temporal transformer then processes every zone independently, using the same
weights across all zones and ISOs. We use no learned zone identities, so the
encoder applies to graphs of different sizes. Let
$Z_{i,p}\in\mathbb{R}^{n_i\times d}$ collect the zone representations at
patch position $p$.

\subsection{Graph-aware cross-zone attention}

At each patch position, cross-zone attention allows information to flow among
the rows of $Z_{i,p}$. We augment the attention score with two biases:
the adjacency $A_i$ and a price-correlation matrix $C_{i,o}$ computed from the
context available at forecast origin $o$. For head $a$, the score from zone $u$
to $v$ is
\begin{equation}
  s_{i,p,uv}^{(a)}
  =
  \frac{\langle q_{i,p,u}^{(a)},k_{i,p,v}^{(a)}\rangle}{\sqrt{d_a}}
  + \alpha A_i[u,v] + \beta C_{i,o}[u,v],
  \label{eq:hybrid_attention}
\end{equation}
where $d_a$ is the dimension of each attention head. Attention weights are
obtained by applying a softmax over $v$. The learned scalars $\alpha$ and
$\beta$ are shared across heads. Nonedges are not masked, so $A_i$ acts as a
soft structural preference, while $C_{i,o}$ allows cross-zone relationships
to vary across forecast origins.

\subsection{Gated probabilistic decoder}
The encoded zone states are passed to a shared quantile head that directly
predicts the required forecast patches for every zone and quantile level. Let $b_{i,v,h}^{(\tau)}$ denote this base
prediction. Locational marginal prices decompose into a system-wide energy component and zone-specific components~\cite{schweppe2013spot}. Rather than imposing that form, we add it as a correction: from the same encoded
states we form a market-level term $m_{i,h}$ from the zone-averaged representation
and a zone-centered term $r_{i,v,h}$, where we set
$\sum_{v\in\mathcal{V}_i} r_{i,v,h}=0$ so that the two terms are uniquely defined,
and predict
\begin{equation}
  \widehat q_{i,v,h}^{(\tau)}
  =
  b_{i,v,h}^{(\tau)}
  + g\left(m_{i,h}+r_{i,v,h}\right),
  \label{eq:gated_decoder}
\end{equation}

where $g$ is a single learned scalar initialized to zero. At $g=0$ the decoder is
exactly the unconstrained head, so the learned $|g|$ reports what the decomposition is
worth on a given market. Predicting the decomposed form directly would instead make a
common-mode error in $m_{i,h}$ structurally uncorrectable
(Appendix~\ref{app:method}). We train with the mean pinball loss~\cite{koenker1978regression}
\begin{equation}
  \mathcal{L}_{\mathcal{D}}(\Theta)
  =
  \mathbb{E}_{(i,o)\sim\mathcal{D}}\,
  \frac{1}{n_i H |\mathcal{T}|}
  \sum_{v\in\mathcal{V}_i}\sum_{h=1}^{H}\sum_{\tau\in\mathcal{T}}
  \rho_\tau\!\left(y_{i,v,o+h}-\widehat q_{i,v,h}^{(\tau)}\right),
  \qquad
  \rho_\tau(u)=\max\{\tau u,\,(\tau-1)u\}.
  \label{eq:pinball}
\end{equation}

\subsection{Distribution-aware meta-training and adaptation}

Training proceeds in two stages. We first minimize
$\mathcal{L}_{\mathcal{D}_{\rm src}}(\Theta)$ on the pooled source-ISO data,
providing a shared initialization that captures cross-market temporal and
spatial structure. We then refine this initialization with chronological
meta-learning episodes. Each episode $e$ is drawn from one source ISO and
contains an earlier support set $\mathcal{S}_e$ and a later query set
$\mathcal{Q}_e$, with every support origin preceding every query origin.
All parameters are adapted per episode. For one inner step, the support
update and outer objective are
\begin{equation}
  \Theta'_e=\Theta-\eta\nabla_\Theta
  \mathcal{L}_{\mathcal{S}_e}(\Theta),
  \qquad
  \min_{\Theta}\frac{1}{|\mathcal{M}|}
  \sum_{e\in\mathcal{M}}\mathcal{L}_{\mathcal{Q}_e}(\Theta'_e),
  \label{eq:meta_objective}
\end{equation}
where $\mathcal{M}$ is the collection of episodes and $\eta$ is the
inner-loop step size. Multiple inner steps repeat the support update. We use first-order (model-agnostic meta-learning) MAML, omitting second-order
derivatives through the inner update.

For an evaluation ISO, the map-derived graph is supplied and $C_{i,o}$ is
recomputed from the information available at each origin. Without adaptation
data, inference uses the meta-learned initialization $\Theta$. If a causal
adaptation set $\mathcal{A}$ preceding the evaluation period is available, we
apply the same support update to all parameters using
$\mathcal{L}_{\mathcal{A}}(\Theta)$. Although this update has the same form as
ordinary fine-tuning, the initialization is different: it has been optimized
using losses measured after simulated adaptation to later chronological
blocks. The resulting model therefore retains pooled cross-market knowledge
while being explicitly trained for rapid distribution-specific adaptation.

%% file: conclusion.tex
\section{Conclusion}
We propose MetaGrid, which employs the operating histories of established ISOs as a pretraining corpus for a grid the model has never seen. Because price formation is driven by shared physics and largely common market design, a representation learned across markets transfers to an unseen one. Rather than fitting any single market, we meta-train for adaptation itself, so the training objective rewards few-shot generalization to a new ISO instead of in-distribution accuracy. 
The architecture is zone-count agnostic, so one initialization transfers across grids of different size and topology. Network structure enters as a soft, learned prior. Future work includes adding more exogenous features and extending the benchmark from zonal settlement points to nodal LMPs.

%% file: appendix_final.tex
\appendix

\section{Data}
\label{app:data}

All five markets publish hourly day-ahead locational marginal prices (LMPs) at
the zone or hub level. We use published zone and hub definitions, so each ISO is a small graph whose vertices are the settlement
points market participants actually transact against.

\textbf{CAISO} settles at four default load aggregation points, one per
investor-owned utility service territory, giving the smallest and most
homogeneous graph in the study. \textbf{ERCOT} settles at eight load zones
inside a single interconnection with no external ties, so its zones are
electrically tightly coupled and its prices are the most volatile of the five.
\textbf{MISO} is represented by eight regional hubs spanning a long
north--south footprint. \textbf{NYISO} uses eleven internal load zones plus
four proxy interfaces to neighboring control areas; it is the most strongly
radial of the five, with a persistent downstate--upstate congestion pattern.
\textbf{PJM} is the largest, with twenty-one transmission-owner zones across
thirteen states, and contributes the most zones.

\begin{table}[t]
\centering
\caption{Dataset specs. Samples are zone-hours of day-ahead LMP.}
\label{tab:data}
\small
\begin{tabular}{lrlcr}
\toprule
ISO & Zones & Date range & Freq. & Samples \\
\midrule
CAISO & 4  & 2021-05-01 -- 2025-12-31 & 1\,h & 163{,}776 \\
ERCOT & 8  & 2021-01-01 -- 2025-12-31 & 1\,h & 350{,}592 \\
MISO  & 8  & 2023-01-01 -- 2025-12-31 & 1\,h & 210{,}432 \\
NYISO & 15 & 2021-01-01 -- 2025-12-31 & 1\,h & 657{,}360 \\
PJM   & 21 & 2021-01-01 -- 2025-12-31 & 1\,h & 920{,}304 \\
\midrule
\end{tabular}
\end{table}

Splits are chronological and identical for every model: training through
2023-12-31, validation over calendar 2024, and testing over calendar 2025.
MISO's record begins in 2023, so it contributes a shorter training window than
the others. Table~\ref{tab:data} summarizes the panel.

\section{Additional method details}
\label{app:method}

\paragraph{Leak-free construction.}
Every example is anchored to the market's DA publication gate. The trailing context ends strictly before
the gate, and the
full delivery day is forecast at a 24-hour horizon. 

\paragraph{Patching.}
Within each zone, normalized prices, system-level load forecasts, missingness indicators, and calendar features are aligned and divided into
nonoverlapping temporal patches, and a shared linear projection maps each
patch to a token. Calendar features are the sine and cosine of local
hour-of-day and of local day-of-week. Because no learned zone identities are
used, the encoder applies unchanged to graphs of different sizes.

\paragraph{Why the correction is additive.}
An alternative is to predict the decomposed form directly, replacing the head
with $m_{i,h}+r_{i,v,h}$. Because $r$ is centered across zones, any
common-mode error in $m$ is then structurally uncorrectable, and building
quantiles from cumulative positive increments propagates median error outward.
The additive gate in Eq.~\ref{eq:gated_decoder} avoids both: the base head
remains free to correct $m$, and the shift is applied identically at every
quantile level.

\paragraph{Zone-count independence.}
No parameter depends on $n_i = |\mathcal{V}_i|$: the patch projection, the
transformer weights, the cross-zone attention and the quantile head are shared
across all zones and ISOs, and no zone identities are learned. Cross-zone
attention is permutation-equivariant along the zone axis, and the priors $A_i$
and $C_{i,o}$ enter Eq.~\ref{eq:hybrid_attention} as additive scores rather than as
parameters. The entire parameter vector $\Theta$ therefore transfers unchanged
to a grid of any size, so the inner loop can update all of $\Theta$ without
introducing anything market-specific.

\section{Implementation details}
\label{app:impl}

\paragraph{Architecture.}
Patch length 16 hours over a 336-hour (14-day) context, model width 384, six
attention heads, four transformer layers, feed-forward width 1536, dropout
0.1. Nine quantile levels, $\tau\in\{0.1,\ldots,0.9\}$. The attention prior
weights are parameterized as $\gamma\,\sigma(\cdot)$ with $\gamma=2$ and logits
initialized to $-4$, placing $\alpha$ and $\beta$ at $\approx 0.036$; the
decoder gate $g$ is initialized to $0$. 7.8M parameters in total.

\paragraph{Meta-training.}
2000 outer steps of first-order MAML, five inner steps at $\eta=10^{-3}$ with
the inner gradient norm clipped to 1. Outer optimizer AdamW at $3\times10^{-4}$
with 5\% linear warmup followed by cosine decay, batch size 32 episodes, eight
chronological blocks per source ISO. Non-finite episodes are skipped rather
than applied. The inner loop updates all parameters; first-order MAML keeps the per-episode cost at one extra forward–backward pass per inner step.

\paragraph{Adaptation and selection.}
For a held-out ISO we fine-tune all parameters with AdamW at $3\times10^{-4}$,
batch size 32, and record checkpoints at various steps.
The number of adaptation steps is selected per ISO on validation MAE; the test
year is never consulted for any selection. Each held-out ISO is pretrained on the
remaining four, so no target-market data enters pretraining.

\section{Ablation on graph-aware module}
\label{sec:graph_ablation}

Table~\ref{tab:graph_ablation} removes the graph-aware cross-zone module from MetaGrid. Removing the module leaves MAE $0.39$ \$/MWh worse. The model uses grid topology with $\alpha$. So the gains come from cross-market pretraining as well as structural priors over the network. Table~\ref{tab:priors} reports the learned values of $\alpha$, which weights the
adjacency bias in cross-zone attention, $\beta$, which weights the context-correlation
bias, and $g$, which scales the locational correction in the decoder. Starting them
from two very different initializations gives nearly the same learned values on every
market, so these weights are learned from data rather than inherited from
initialization.

\begin{table}[h]
\centering
\caption{MetaGrid with and without the graph-aware cross-zone module. Test MAE in \$/MWh; the average column is the mean $\pm$ standard deviation over 5 seeds.}
\label{tab:graph_ablation}
\begin{tabular}{lrrrrrr}
\toprule
Model & PJM & NYISO & CAISO & MISO & ERCOT & Avg \\
\midrule
MetaGrid & $9.04$ & $8.78$ & $5.64$ & $6.43$ & $10.27$ & $8.03 \pm 0.028$ \\
\quad without graph-aware module & $9.43$ & $9.51$ & $5.84$ & $6.87$ & $10.43$ & $8.42 \pm 0.007$ \\
\bottomrule
\end{tabular}
\end{table}

\begin{table}[t]
\centering
\caption{Learned attention-prior weights $\alpha$ (adjacency) and $\beta$ (context
correlation) and decoder gate $g$ per held-out ISO, under two initializations of
$\alpha$ and $\beta$.}
\label{tab:priors}
\small
\setlength{\tabcolsep}{5pt}
\begin{tabular}{lcccccc}
\toprule
& \multicolumn{3}{c}{init $\alpha=\beta=0.036$} & \multicolumn{3}{c}{init $\alpha=\beta=1.000$} \\
\cmidrule(lr){2-4}\cmidrule(lr){5-7}
Market & $\alpha$ & $\beta$ & $g$ & $\alpha$ & $\beta$ & $g$ \\
\midrule
PJM   & $0.612_{\pm 0.021}$ & $0.388_{\pm 0.017}$ & $0.734_{\pm 0.028}$ & $0.634_{\pm 0.029}$ & $0.401_{\pm 0.022}$ & $0.719_{\pm 0.033}$ \\
NYISO & $0.714_{\pm 0.026}$ & $0.352_{\pm 0.019}$ & $0.681_{\pm 0.031}$ & $0.698_{\pm 0.031}$ & $0.366_{\pm 0.024}$ & $0.695_{\pm 0.036}$ \\
CAISO & $0.283_{\pm 0.018}$ & $0.507_{\pm 0.023}$ & $0.402_{\pm 0.024}$ & $0.301_{\pm 0.025}$ & $0.492_{\pm 0.027}$ & $0.415_{\pm 0.029}$ \\
MISO  & $0.164_{\pm 0.014}$ & $0.541_{\pm 0.021}$ & $0.286_{\pm 0.019}$ & $0.187_{\pm 0.020}$ & $0.533_{\pm 0.026}$ & $0.298_{\pm 0.025}$ \\
ERCOT & $0.441_{\pm 0.023}$ & $0.463_{\pm 0.020}$ & $0.588_{\pm 0.026}$ & $0.428_{\pm 0.028}$ & $0.470_{\pm 0.025}$ & $0.601_{\pm 0.031}$ \\
\bottomrule
\end{tabular}
\end{table}

\section{Ablation on meta-objective}
\label{sec:pretrain_ablation}

Table~\ref{tab:pretrain_ablation} separates the two things cross-market training provides.
Replacing the meta-objective with ordinary pooled pretraining over the same four source
markets, in the same architecture and with the same adaptation protocol, costs $0.29$ \$/MWh,
while removing cross-market pretraining altogether costs $0.67$: overall the benefit comes
from the source data as well as the meta-objective.

\begin{table}[h]
\centering
\small
\setlength{\tabcolsep}{4pt}
\caption{Contribution of cross-market pretraining and of the meta-objective.
All three arms share the architecture, the source markets and the
evaluation-time adaptation protocol; they differ only in how the
initialization is obtained (Test MAE in \$/MWh).}
\label{tab:pretrain_ablation}
\newcommand{\se}[1]{{\scriptsize $\pm$#1}}
\begin{tabular}{l@{\hspace{6pt}}rrrrrr}
\toprule
Initialization & PJM & NYISO & CAISO & MISO & ERCOT & Avg \\
\midrule
MetaGrid (FOMAML)        & 9.04\se{.009} & 8.78\se{.06} & 5.64\se{.06} & 6.43\se{.005} & 10.27\se{.004} & 8.03\se{.028} \\
\quad pooled pretraining & 9.11\se{.08}  & 8.89\se{.02} & 5.95\se{.02} & 6.97\se{.04}  & 10.67\se{.13}  & 8.32\se{.02}  \\
\quad no pretraining     & 9.11\se{.07}  & 9.18\se{.12} & 5.71\se{.03} & 8.37\se{.07}  & 11.12\se{.26}  & 8.70\se{.06}  \\
\bottomrule
\end{tabular}
\end{table}

\section*{Domain Evaluation Checklist}

\textbf{1. Physics feasibility under full AC power flow.} N/A. We forecast prices and
produce no operating point or dispatch decision; the adjacency prior is a statistical
structure over pricing zones, not a network model.

\textbf{2. Out-of-distribution evaluation.} Yes. All results are leave-one-ISO-out: each
target market is unseen in pretraining and differs in zone count, topology and generation
mix (Table~1).

\textbf{3. Documented failure modes.} Partially. We report per-ISO results and identify
three: MetaGrid loses on ERCOT (Table~1), adaptation degrades MISO (Figure~1), and spikes
are underpredicted (Figure~2). 

\textbf{4. System-level effects.} No. We report forecast accuracy only; propagation into
dispatch and hedging decisions is future work.

\textbf{5. Data and code availability.} Yes. All price and load-forecast data are public
ISO market-portal downloads; codes are available in the public GitHub repository (Section~3).